\documentclass{article}

\usepackage{PRIMEarxiv}
\usepackage{amsmath}
\usepackage[utf8]{inputenc} 
\usepackage[T1]{fontenc}    
\usepackage{hyperref}       
\usepackage{url}            
\usepackage{booktabs}       
\usepackage{amsfonts}       
\usepackage{nicefrac}       
\usepackage{microtype}      
\usepackage{lipsum}
\usepackage{fancyhdr}       
\usepackage{graphicx}       
\graphicspath{{images/}} 
\usepackage{amssymb}
\usepackage{algorithm}
\usepackage{algpseudocode}
\usepackage{multirow}
\usepackage{natbib}
\usepackage[section]{placeins}
\usepackage{tablefootnote}
\usepackage{flafter} 
\usepackage[bottom]{footmisc}

\usepackage{graphicx}

\usepackage{lipsum}

\newcommand\blfootnote[1]{%
  \begingroup
  \renewcommand\thefootnote{}\footnote{#1}%
  \addtocounter{footnote}{-1}%
  \endgroup
}

\algrenewcommand\algorithmicrequire{\textbf{Input:}}
\algrenewcommand\algorithmicensure{\textbf{Output:}}

\title{Counterfactually Fair Regression with Double Machine Learning
}

\author{
  Patrick Rehill \\
  Centre for Social Research and Methods \\
  Australian National University \\
  Canberra\\
  \texttt{patrick.rehill@anu.edu.au} \\
}

\begin{document}
\maketitle

\begin{abstract}
Counterfactual fairness is an approach to AI fairness that tries to make decisions based on the outcomes that an individual with some kind of sensitive status would have had without this status. This paper proposes Double Machine Learning (DML) Fairness which analogises this problem of counterfactual fairness in regression problems to that of estimating counterfactual outcomes in causal inference under the Potential Outcomes framework. It uses arbitrary machine learning methods to partial out the effect of sensitive variables on nonsensitive variables and outcomes. Assuming that the effects of the two sets of variables are additively separable, outcomes will be approximately equalised and individual-level outcomes will be counterfactually fair. This paper demonstrates the approach in a simulation study pertaining to discrimination in workplace hiring and an application on real data estimating the GPAs of law school students. It then discusses when it is appropriate to apply such a method to problems of real-world discrimination where constructs are conceptually complex and finally, whether DML Fairness can achieve justice in these settings.
\end{abstract}

\keywords{AI fairness \and double machine learning \and fair regression \and counterfactual fairness}

\section{Introduction}
\blfootnote{\textbf{Statements and Declarations:} No funding was received for conducting this study. The author has no relevant financial or non-financial interests to disclose.}Machine learning systems are being used more and more to make (or help make) decisions in high-stakes domains like medicine, law enforcement and hiring. This brings with it concerns about ensuring these systems are fair. Exactly what fair means differs across different sources, but the general idea is that there are certain bases for decision-making that would be unjust if a human decision-maker were to use them (e.g. race, gender) and we should therefore make sure our algorithms are not using these for their predictions and decisions as well. It is generally not enough to just guard against discrimination on these sensitive constructs by omitting measurements of them as predictors \citep{mehrabi_survey_2019}.\footnote{In general, the literature on AI Fairness sometimes fails to make the distinction between constructs and the variables that represent them. This is a more commonly discussed topic in the social sciences where measurement of a construct is often quite challenging. This distinction will be important for discussing the limitations of the counterfactual approach to fairness so rather than using the terms often used for the things that are protected by AI fairness like ‘sensitive attributes’ or ‘protected classes’, this paper uses the terms ‘sensitive constructs’ and ‘sensitive variables’.} Black box models often find ways to proxy these constructs based on seemingly innocuous variables that end up correlating with them \citep{pedreshi_discrimination-aware_2008}. For example, in 2014 Amazon began trying to develop a machine learning tool to predict which potential hires would have the most success at the company. However, the project was abandoned the next year because the tool discriminated against women. The reason for this is that due to historic gender inequalities in the sector, the algorithm had identified in the training data that successful employees tended to not have gone to a women’s college or been captain of a women’s sport team \citep{dastin_amazon_2018}. The problem of making models that do not discriminate based on sensitive attributes is a challenging one and a large literature on AI fairness has emerged to try and address it.

There are many ways to measure and address fairness concerns in a machine learning model and it can be hard to know which to use (comprehensive reviews of the literature on different metrics of fairness can be found in \citet{mehrabi_survey_2019} and \citet{corbett-davies_measure_2018}). Complicating matters further, even a handful of the most common metrics are not jointly satisfiable \citep{hedden_statistical_2021}, i.e. trying to achieve one measure of fairness often prevents us from achieving others. Most of the commonly used fairness metrics are based on group outcomes, for example whether men and women have the same odds of receiving a beneficial decision. A very different approach is that of counterfactual fairness which seeks to measure fairness at the individual level by asking whether someone would have received the same outcome in a counterfactual world where they did not belong to one or more protected classes. \citet{kusner_counterfactual_2018} formally defines counterfactual fairness as for predictor $\hat{Y}$, for any set of sensitive variables $D=d$ and non-sensitive variables $X=x$

$$P\left({\hat{Y}}_{D\gets d}=y\mid X=x,D=d\right)=P\left({\hat{Y}}_{D\gets d^\prime}=y\mid X=x,D=d\right)$$ for all $y$ and for any possible value of $d^\prime$.\footnote{The is not an exact reproduction of Kusner et al.'s definition as this tweaks the original definition to bring it in line with notation used later in this paper.}

Counterfactual fairness offers a range of tools for thinking through AI fairness including the use of causal inference methods as essentially counterfactual reasoning is causal reasoning \citep{kusner_counterfactual_2018, kasirzadeh_use_2021}. With good causal inference, one can be confident that the model is counterfactually fair with no need to verify this through observable fairness metrics; although there is also no way to directly measure this performance per the Fundamental Problem of Causal Inference which says that in real-world settings, counterfactual outcomes are impossible to observe \citep{holland_statistics_1986}.

While statistically, causal inference techniques should lead to fair outcomes at the individual level, it is important to also consider critiques about whether causal inference can actually meaningfully conceive of counterfactuals for complex constructs like race or gender \citep{kasirzadeh_use_2021}. An important theme that will recur throughout this paper is that if a variable is problematic enough that it needs protecting, it is probably also difficult to define what one means to measure with this variable, how best to measure it, and how one should imagine a counterfactual that manipulates it \citep{hanna_towards_2020, kasirzadeh_use_2021}. On top of this, there are arguments to be had about whether just protecting variables counterfactually without ensuring some parity in outcomes is fair (an argument the author is sympathetic to). As any causal inference scholar could attest, it is very hard to get causal inference from observational data right \citep{imbens_causal_2015}, it may lead to unfair outcomes and it is difficult to even know this given the ground-truth is not observable. In short, counterfactual fairness is powerful in theory, but involves difficult assumptions that should be considered properly.

The vast majority of AI fairness approaches are focused on classification problems, however in this paper, our focus will be on regression as the approach taken to partialling out the effect of sensitive variables relies on least squares. Some papers have already tackled the problem of fairness in regression but take different approaches \citep{agarwal_fair_2019,berk_convex_2017,steinberg_fast_2020,komiyama_nonconvex_2018}. The fact that the fair regression literature is so small speaks to the importance of classification problems in modern machine learning, but also to the difficulty that comes with applying criteria of fairness to a result without discrete outcomes (and a small number of them) that are easy to classify as positive or negative.

This paper proposes a new approach to mitigate discrimination based on work in the causal machine learning literature, in particular the Neyman-orthogonality approach from \citet{chernozhukov_doubledebiased_2018}. This approach removes the effect of a set of biasing variables $D$ from outcomes and predictors in order to give unbiased predictions. Assuming the effects of sensitive and non-sensitive variables are additively separable with respect to the target variable (i.e. sensitive and non-sensitive variables may interact within these categories but not between them) it uses Chernozhukov et al.’s adaptation of The Frisch-Waugh-Lovell (FWL) Theorem \citep{frisch_partial_1933,lovell_seasonal_1963} to employ machine learning methods which partial out the effect of the sensitive variables on other variables. Then, rather than fitting a causal model in the final stage like Chernozhukov et al., it fits an arbitrary machine learning model to make fair predictions of outcomes. 

Our method can be seen as a specific case of Kusner et al.'s approach (in particular their additive error approach) which allows for the use of arbitrary machine learning methods in the regression case. While Kusner et al. do not rule out the use of more complex machine learning methods, their additive error examples rely on linear models because of the ease with which one can partial out effects in a way that protects the error term. The use of double machine learning allows for a model that can fit more complex functions with machine learning methods just as sophisticated as the final prediction algorithm under theoretical guarantees that it will not over-fit or under-fit the effect of these variables. This is the property that makes DML causal estimates unbiased and $\sqrt{N}$ consistent \citep{chernozhukov_doubledebiased_2018} and analogously gives counterfactual outcomes the same property as well. That is to say, with DML one can be confident in protecting the error distribution for arbitrary regression methods just as one could be in the linear case. Even though the treatment effect is not a target of this analysis, a counterfactual estimate can be found using the process of causal estimation. As per the potential outcomes framework \citep{imbens_causal_2015}, a treatment effect ($\tau$) in the binary case is the difference in potential outcomes $Y$ of different treatment statuses ($W$).  $$\tau=Y(W=1) - Y(W=0)$$As the counterfactual outcome is never observed, it can be estimated in a simple binary case in terms of an observed outcome and treatment effect as $$\hat{Y_{cf}}=Y_{obs}+(1-2w)\hat{\tau}$$
This paper uses this relationship to transform the problem of fairness into one solvable by econometric methods.

The rest of this paper is structured as such. Section 2 introduces the Potential Outcomes framework and how it can be used in counterfactual fairness. Section 3 defines the DML Fairness method. Section 4 discusses the assumption of additive seperability in the outcome function and why this assumption is useful. Section 5 consists of a simulation study. Section 6 consists of an application of the method to real-world data. Finally, Section 7 discusses the prospects for these methods and its drawbacks. While DML Fairness could be a powerful tool for achieve fair regression, it is not a panacea to be applied without thought to the context in which it is being used.

\section{Potential Outcomes and counterfactual fairness}
\label{sec:bg}

The counterfactual fairness literature generally uses the structural causal modelling (SCM) approach to causal inference laid out by Pearl, Glymour and others \citep{pearl_causality_2009,pearl_bayesianism_2001,spirtes_causation_2001,glymour_causation_2009}. The \citet{kusner_counterfactual_2018} approach does not neccessarily involve a full structural causal model, but if the model includes descendents of sensitive variables (i.e. variables affected by sensitive variables), a causal graph is necessary. The use of an SCM approach makes sense given it is the dominant one in computer science where AI fairness work tends to be done \citep{imbens_potential_2020}. However, the rival Potential Outcomes (PO) framework of causal inference \citep{imbens_causal_2015} is formally equivalent \citep{galles_axiomatic_1998} and has powerful tools for estimating causal effects which in the process estimate counterfactuals. A PO-based approach does not require explicitly defining a causal model and may be more useful in high-dimensional datasets where it would be difficult to define a full causal model. In this case one can instead make the assumption that these variables are not the causal descendants of other variables in the dataset. This is often the case as protected characteristics are generally ‘pre-treatment’ variables not causally affected by other variables that might be in our model for example, race and gender are largely immutable constructs that are unlikely to change as a result of other variables. This is no coincidence, the immutability of these characteristics is an important part of why they are considered sensitive in the first place \citep{clarke_against_2015}. In cases where protected variables may in fact be descended from other non-protected variables, the counterfactual can still be estimated, though here more specific assumptions are needed. For example, a sensitive variable like marital status could be a descendent of some nonsensitive variables and an ancestor of others. Here the model should only partial the variable's effect in cases where it is an ancestor. This would involve making the kinds of explicit causal assumptions needed for mediated effect analysis under the PO framework.

The key benefit of relying on the PO framework though is that it allows us to make use of DML. DML allows for the partialling out of effects using the same kinds of methods one would already be using for predictive machine learning rather than the parametric approaches that are the norm in causal inference \citep{breiman_statistical_2001,daoud_statistical_2020}. The problem is that predictive machine learning methods give biased estimates of the effects of individual variables. In order to improve predictive fit in unseen data, most complex machine learning models regularise variable effects. \citep{chernozhukov_doubledebiased_2018} gets around the problem of regularisation bias using a method adapted from the FWL theorem to partial out confounder effects and get an unbiased estimate. In the first stage, it regresses confounders (in this case, sensitive variables) on the treatment variable (the nonsensitive predictor variables) and outcome. Then in the second stage, it makes an estimate of treatment effect by assuming a parametric functional form for the unconfounded effect and then fitting a parametric model for inference. In our approach we map the idea of confounding to sensitive variables, and we skip the inference instead fitting a predictive model directly on the residuals. This is a similar approach to that taken by R-Learner to learn heterogeneous treatment effects with the second-stage model \citep{nie_quasi-oracle_2020}, however in this case, the final model is not being used for causal inference, but instead for prediction. This means there is no need to develop a new loss function, to proxy causal effect, one can test predictions against residualised outcomes in training and use observed outcomes to estimate real-world performance.

\section{Counterfactual DML Fairness}
\label{sec:dmlfairness}
DML Fairness partials out the effect of a set of sensitive variables $D$ from nonsensitive variables $X$. In order to use Neyman orthogonal DML one must assume that the effect of $D$ is additively separable from that of $X$, i.e. the underlying data-generating process is:
$$Y_i=f\left(X_i\right)+g_Y\left(D_i\right)+\varepsilon_{Yi}$$
$$X_{ji}=g_{X_j}\left(D_i\right)+\varepsilon_{X_ji}$$
Importantly, as this is not causal inference, there is no unconfoundedness assumption needed. Rather one only need to 'control' for any variables that one is interested in protecting. Per Chernozhukov et al., DML makes a machine learning estimate of $\hat{Y}$ and each $\widehat{X_j}$ as a function of $D$. While Chernozhukov et al. partials out the effect of a set of adjustment variables on one treatment variable, it is trivial to apply this logic of partialling out to a number of independent variables as previously discussed (Lovell 1963).
$$\widehat{Y_i}=\widehat{g_Y}\left(D_i\right)+\varepsilon_{Yi}$$
$$\widehat{X_{ji}}=\widehat{g_{X_j}}\left(D_i\right)+\varepsilon_{X_ji}$$
These values can then be residualised by taking $Y-\hat{Y}$ and $X-\hat{X}$ to get $\widetilde{X}$ and $\widetilde{Y}$. This rendering the non-sensitive predictors Neyman-orthogonal to the sensitive variables \citep{chernozhukov_doubledebiased_2018}. Just as in DML, cross-fit estimators are necessary to prevent overfitting of the nuisance models which threaten the consistency of the counterfactual estimate.
The final stage fits:
$$\hat{f}_{\widetilde{Y}}(\widetilde{X} )+\varepsilon_{Yi}={\hat{Y}}_{DMLfair}$$
Note that the error term for this residual regression is identical to that in the data generating process which included the biasing sensitive variables. A simple proof for this can be found in \citet{lovell_simple_2008}.
From here an arbitrary predictive method can estimate fair outcomes. These residuals should then give significantly fairer predictions (assuming the nuisance models perform well) than a model estimating outcomes just from the original non-protected variables.

While the regression results lead to fair relative outcomes, the actual outcomes will be very different from the outcomes in the training data. This is because the predictions will be made based on rescaled variables because they only include the portion of variation orthogonal to sensitive predictors. This might be acceptable for some applications where only relative predictions are necessary, but for others it is possible to re-centre the predictions such that they make sense to anyone looking to interpret them. Here the predicted outcomes for a constant set of values for D can be added to the orthogonal portion to yield a recentred estimate. We redefine ${\hat{Y}}_{DMLfair}$ as:
$${\hat{Y}}_{DMLfair}=\widetilde{Y_i}+\widehat{g_Y\left(D_{BC}\right)}=f({\widetilde{X}}_i)$$
This uses the $D$ variables from a defined ‘base case’. This is the counterfactual in terms of which we want to express our outcome. For example, we might want to express predictions in terms of a set of non-marginalised characteristics or the median set of characteristics. Because $\widehat{g_Y\left(D_{BC}\right)}$ is a constant, the actual choice of this value does not affect the fairness implications of the predictions. This is a property of the model being non-interactive between the sensitive and non-sensitive attributes with regards to the outcome. 

DML Fairness can either be used in pre-processing data or in cases where it is useful to trade-off fairness for predictive performance on real data it could be used as a regularliser where the objective function is trying to minimise loss on DML fair and raw data simultaneously with a tuning parameter $\lambda$ controlling the trade-off between fairness and accuracy on observed outcomes. 
$$\hat{f}(X)={\rm argmin}_{\theta}{\left[(1-\lambda)\mathcal{L}(Y,m_Y(X;\theta))+\lambda\mathcal{L}(\widetilde{Y},m_{\widetilde{Y}}(\widetilde{X};\theta))\right]}$$
\section{The additive separability assumption}
\label{sec:addsep}

While there are DML approaches that can accommodate interaction between sensitive and non-sensitive variables \citep{chernozhukov_doubledebiased_2018,colangelo_double_2021} and which therefore could potentially form the basis for a more general DML Fairness approach, this paper covers only the additively separable case. This is because this simplifies the approach greatly and also brings with it useful guarantees. As previously discussed, because of this assumption it is easy to add a constant base case to prediction residuals to recentre the distribution of predictions. It also grants a kind of group-level fairness where for any set of sensitive variables $D=d$ where $\widehat{g_Y}(D)$ is correctly specified, $$E\left[Y_{DMLfair}|D=d\right]\simeq E\left[Y_{DMLfair}\right] \because \widetilde{Y},\widetilde{X} \perp D$$ This means that expected outcomes are approximately equal across groups and across individuals conditional on $\widetilde{X}$. 

Finally, it is much easier to troubleshoot problems in the functioning of the method if the effects of sensitive and nonsensitive variables are easy to separate. Making this assumption though raises two related questions, whether specifically equalising outcomes is counterfactually fair and whether the assumption of additive separability is a realistic one. While AI fairness metrics are largely committed to equalising outcomes and simply differ on which statistics to equalise, counterfactual fairness approaches in theory account for the full range of individual factors that might bias outcomes. Under this idealised counterfactual fairness, there is no need for group-level guarantees and they would in fact bias estimates \citep{dwork_fairness_2012}. The problem is that we do not live in a  world where we have access to ground-truth counterfactual outcomes. Because of this, approximate group-level fairness is a good guide to make the model isn't badly misspecified and does not therefore lead to grossly unjust outcomes.

This is not an unreasonable assumption to make. It is assumed that these sensitive constructs are in the language of Pearl, d-separated from the non-sensitive constructs (e.g. latent ability) on which we wish to predict (which implies additive separability of the terms). While the actual variables themselves will likely not be d-separated from each other (otherwise there would not be much need for a fairness algorithm), recovering these latent constructs is the whole aim of the method and of the \citet{kusner_counterfactual_2018} approach more broadly. To quibble with this assumption of d-separation, one has to make a case that the d-separation of the sensitive and non-sensitive constructs does not exist and to the extent that it does, we are confident in discriminating on this basis. This question of confidence is important because in the kinds of complex social models where we call for AI fairness, it is very difficult to know for sure that assumptions about the underlying DGP are true and that we are measuring constructs correctly.

The history of social science (and biology) is filled with examples of research about differing abilities across different characteristics which we might call sensitive variables. These analyses have been used to reinforce existing power structures and attempts to show differences in abilities by race or gender have often employed poor measurement and confused innate characteristics with socially constructed ones that have been defined by these same power structures \citep{belkhir_race_1994,saini_superior_2019}. As builders of AI systems we should be humble and understand this past (not to mention the past of predictive AI systems being used opaquely for these same purposes \citep{oneil_weapons_2016}). By assuming group-level equality of latent factors we are choosing to assume group differences are essentially errors in measurement or modelling. We are assuming that there are not innate differences across values of sensitive variables like race or gender, and that it is not valuable to discriminate on the basis of these. In the context of the history of statistics, this seems like the safest assumption.

\section{Simulation}
\label{sec:simulation}
For this simulation, we imagine an organisation in an industry historically dominated by white men that is hiring in recent graduates and wants to build a model to predict which applicants are likely to succeed. The model is trained to predict a manager’s rating of a graduate’s job performance after one year based on their university average mark and a problem-solving task in the interview. The organisation currently lacks diversity and management have realised that this is due to employees who are white and men being more likely be scored highly on the problem solving task as this task is one more culturally familiar to this in group and there is a subjective component in the interviewer's assessment of the task. They are also more likely to score well on their performance review (and therefore be retained on staff) once their graduate year is over because those running this review are more likely to positively assess those similar to themselves. In fact, due to historically not being a part of the field, people who are not white or not men are also less likely to score well at university as they are less likely to be able to build peer and mentor relationships. The organisation wants to adjust for these factors in their hiring decisions to make the process fairer, build a more diverse workforce and hopefully find candidates with potential who would have otherwise been passed over.

We generate simulated data according to the procedure in Table 1. To help with this we used the Wakefield package \citep{rinker_wakefield_2018} for R. We generate a dataset of 7000 i.i.d. sampled cases. 5000 cases will be used in the training set and 2000 in the test set. Where variables were categorical and not binary, we coded dummy variables for use in the analysis. The data-generating process is laid out in Table \ref{tab:simdgp}.
\begin{table}[!htp] 
\begin{center}
\begin{tabular}{|c c |} 
 \hline
  Variable & Generation \\
 \hline
 \hline
 \textit{Random variables} & \\
 \hline
 Ability (latent) & Normal distribution with mean = 88 and sd = 4\\
 \hline
 Age & Discrete uniform distribution from 18 to 25\\
 \hline
 Gender & \shortstack{Discrete distribution with probabilities:
	\\ p(gender\ =\ male)\ =\ 0.7
	\\ p(gender\ =\ female)\ =\ 0.275
	\\ p(gender\ =\ nonbinary)\ =\ 0.025
 }\\
 \hline
 Race & \shortstack{Discrete distribution with probabilities approximately \\matching US racial make-up (see \citet{rinker_wakefield_2018}})\\
 \hline
 \hline
 \textit{Dependent variables} & \\
 \hline
 Hiring problem-solving assessment & \shortstack{assessment = 0.1(ability)$\cdot$(gender\_male+1.07)$\cdot$\\(race\_white+1.07)+$\varepsilon_{assessment}$ }\\
 \hline
 Grade average & grade=0.1(ability)$\cdot$(gender\_male+1.29)+$\varepsilon_{grade}$\\
 \hline
 Performance review score & \shortstack{
 rating = 3.1(sin(age)+3.7(gender\_male)+\\1.9(race\_white)+7(gender\_male$\cdot$race\_white)+\\0.7(grade)+0.13(assessment)+$\varepsilon_{rating}$
 }\\
 \hline
 \hline
 \textit{Error terms} & \\
 \hline
 Hiring problem-solving assessment error ($\varepsilon_{assessment}$)	& Normal distribution with mean = 0 and sd = 1.09\\
 \hline
Grades error ($\varepsilon_{grade}$) & Normal distribution with mean = 0 and sd = 1.09\\
\hline
Performance review rating error ($\varepsilon_{rating}$) &Normal distribution with mean = 0 and sd = 10\\
 \hline
\end{tabular}
\caption{\label{tab:simdgp}Data-generating process for simulation study}
\end{center}
\end{table}
\FloatBarrier
Note that the discrimination effects were defined simply as white or not and male or not. This is obviously simplistic, but given that there are relatively large categories (e.g. women or Hispanics) and relatively small categories (e.g. non-binary people and Hawaiians) with the same effect size, this will give a sense of how important a large sample size is for achieving good counterfactual estimates.
We treat age, gender and race as sensitive variables and grades and problem-solving assessment as non-sensitive predictors. We use random forest regressors implemented in the ranger package \citep{wright_ranger_2017} for R with 2000 trees for each model. While small random forests can fit linear models poorly \citep{breiman_random_2001}, it is a good ‘general-purpose’ algorithm for making causal inferences on tabular data \citep{mcconnell_estimating_2019}. Assuming we do not know the underlying data-generating process is linear for some effects (for example, the performance review rating once orthogonalised to sensitive variables), the random forest is a good choice. We added the predictions from the residuals to the base case score of an 18 year-old white man.
For training, we use a cross-fitting procedure with ten folds where the nuisance models for 10\% of cases are estimated based on a model fit on the remaining 90\% of cases. This reduces overfitting substantially in nuisance estimates. These are the values that are then used to train the predictive model on the residuals.
When it comes to estimating nuisance values for the held-out test data, we use an average of the 10 models to get an estimate of each nuisance parameter. We then use these parameters to get the out-of-sample goodness of fit of the predictive estimate.

\begin{figure}[!htp] 
  \includegraphics[scale = 0.6]{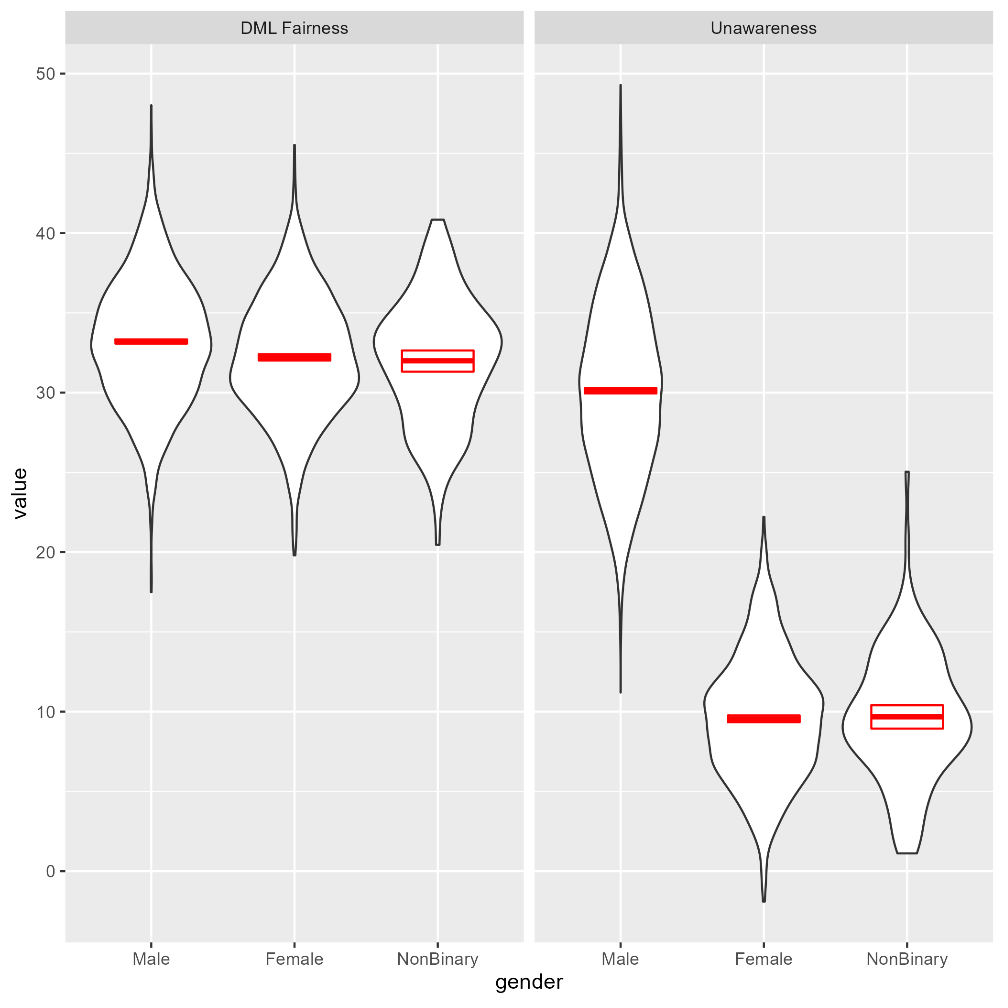}
  \caption{Comparison of DML fair and fairness through unawareness estimates by gender (with mean estimate and 95\% confidence intervals in red)}
  \label{fig:fig1}
\end{figure}

\begin{figure}[!htp] 
  \includegraphics[scale = 0.6]{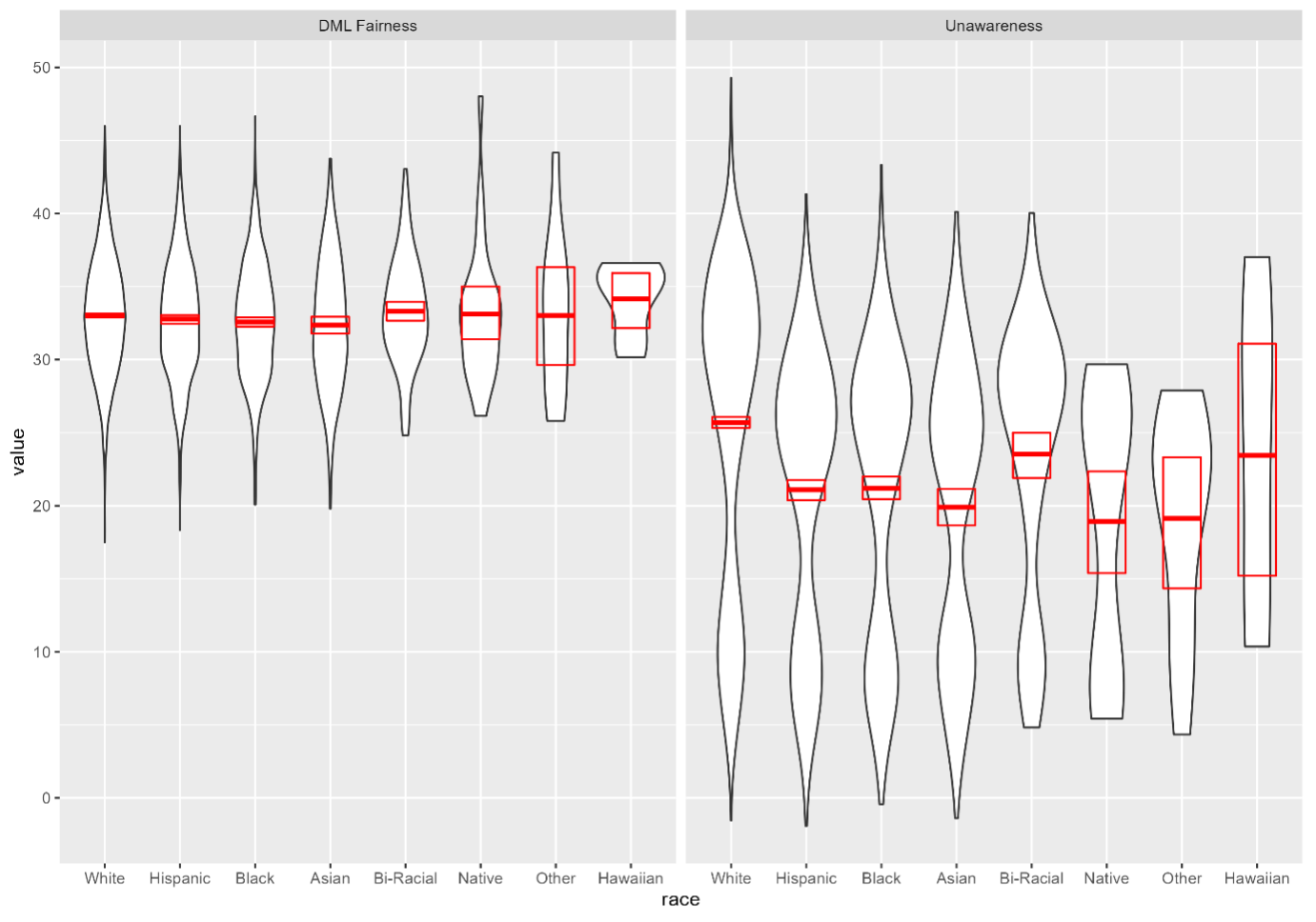}
  \caption{Comparison of fair and unfair estimates by race (with mean estimate and 95\% confidence intervals in red) }
  \label{fig:fig2}
\end{figure}

\begin{figure}[!htp] 
  \includegraphics[scale = 0.6]{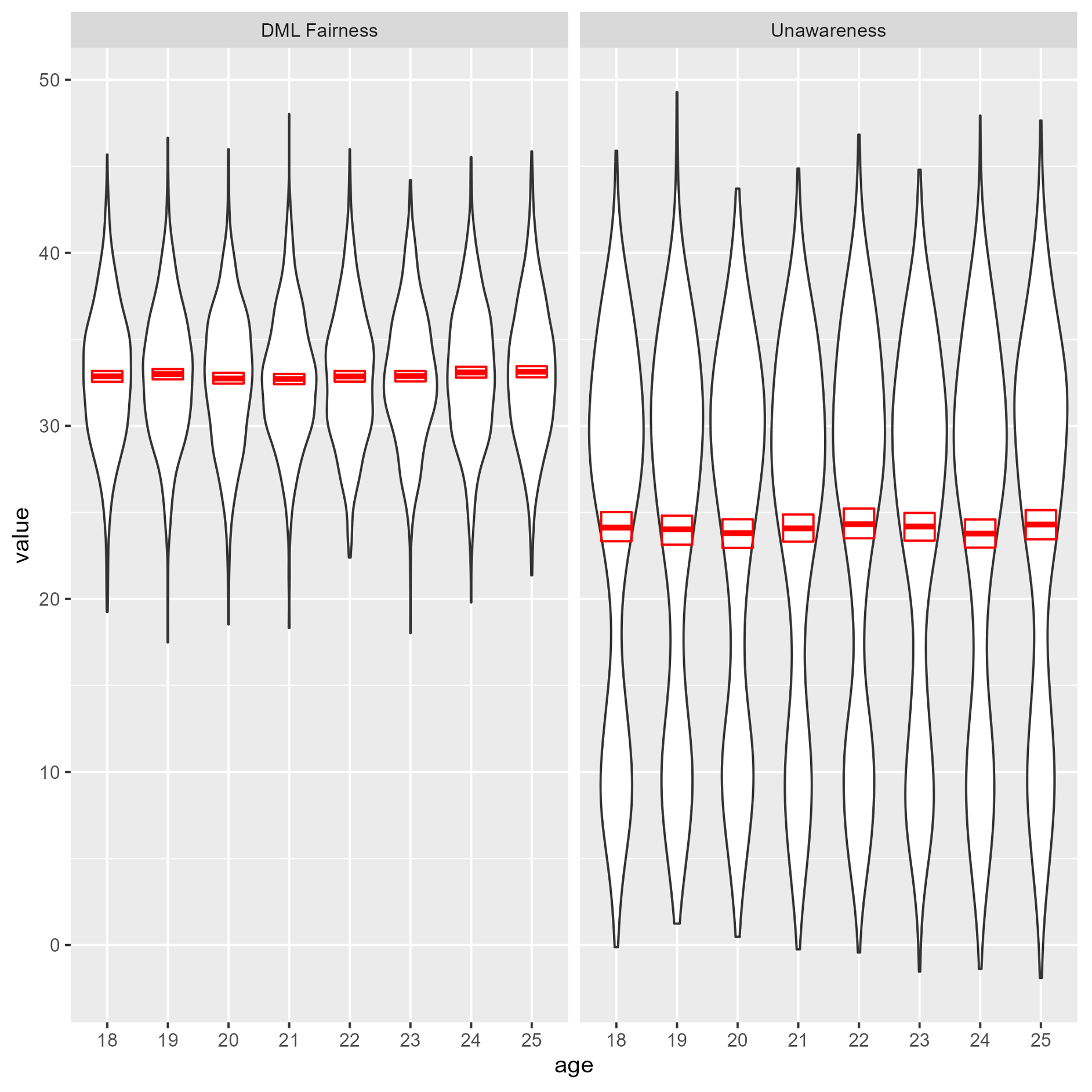}
  \caption{Comparison of fair and unfair estimates by age (with average ability in red)}
  \label{fig:fig3}
\end{figure}

\begin{figure}[!htp] 
  \includegraphics[scale = 0.6]{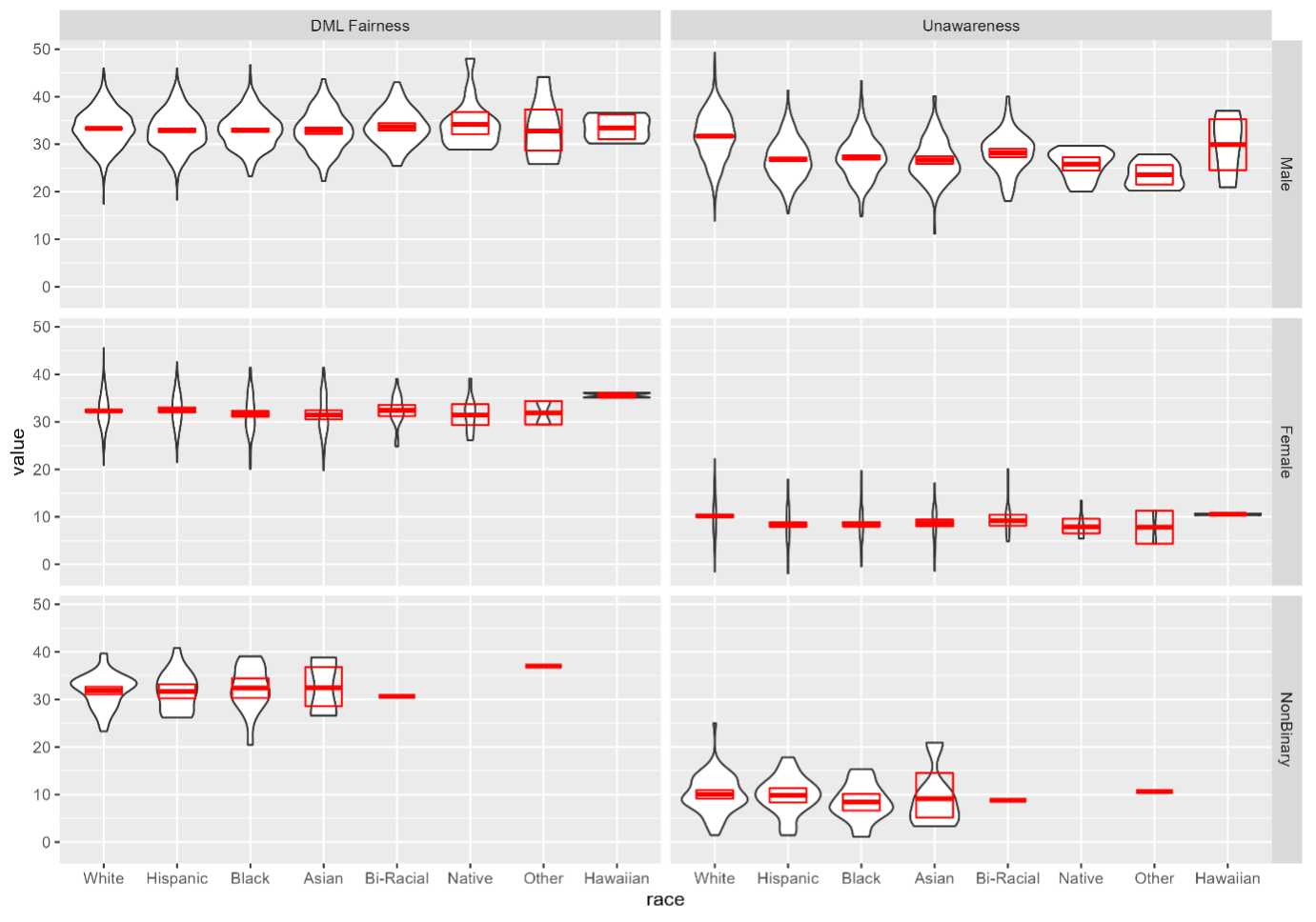}
  \caption{Comparison of fair and unfair estimates by race and gender (with average ability in red)}
  \label{fig:fig4}
\end{figure}
\begin{figure}[!htp] 
  \includegraphics[scale = 0.6]{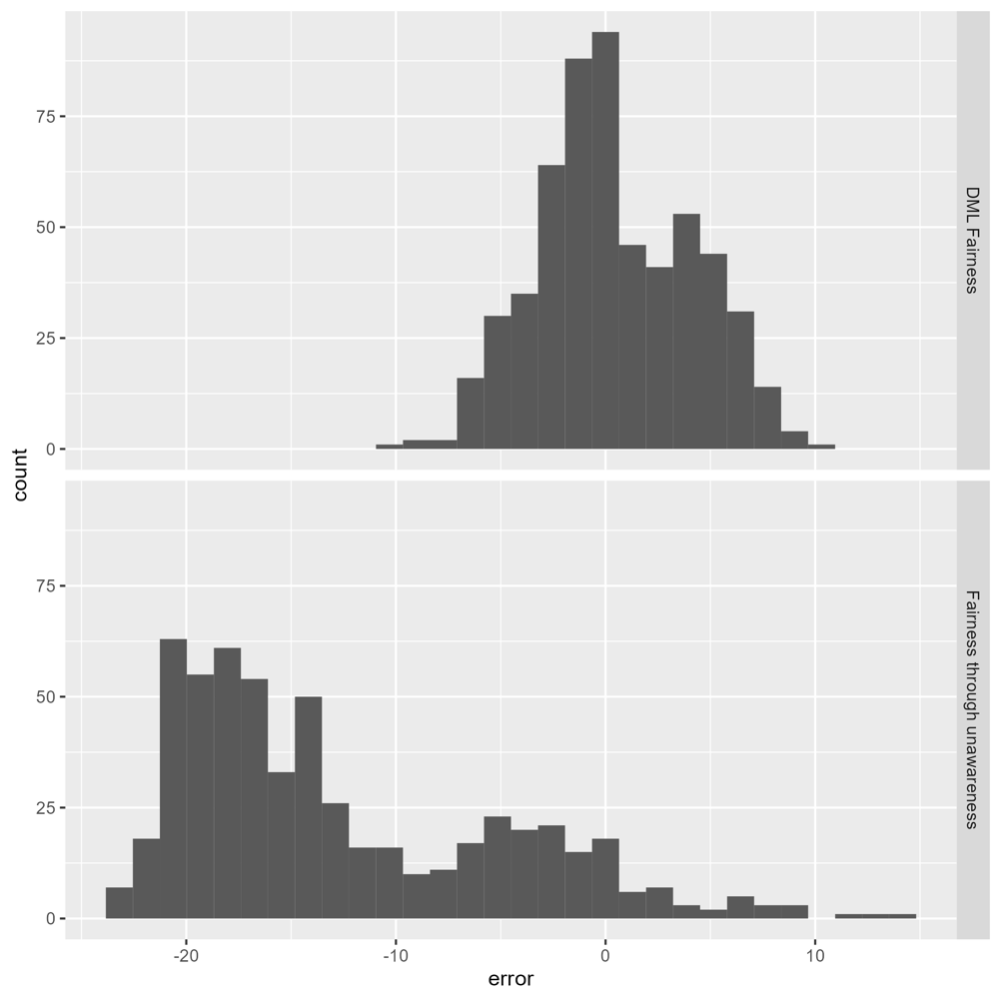}
  \caption{Counterfactual error for non-white, non-men manipulated to be white men}
  \label{fig:fig5}
\end{figure}
\begin{figure}[!htp] 
  \includegraphics[scale = 0.6]{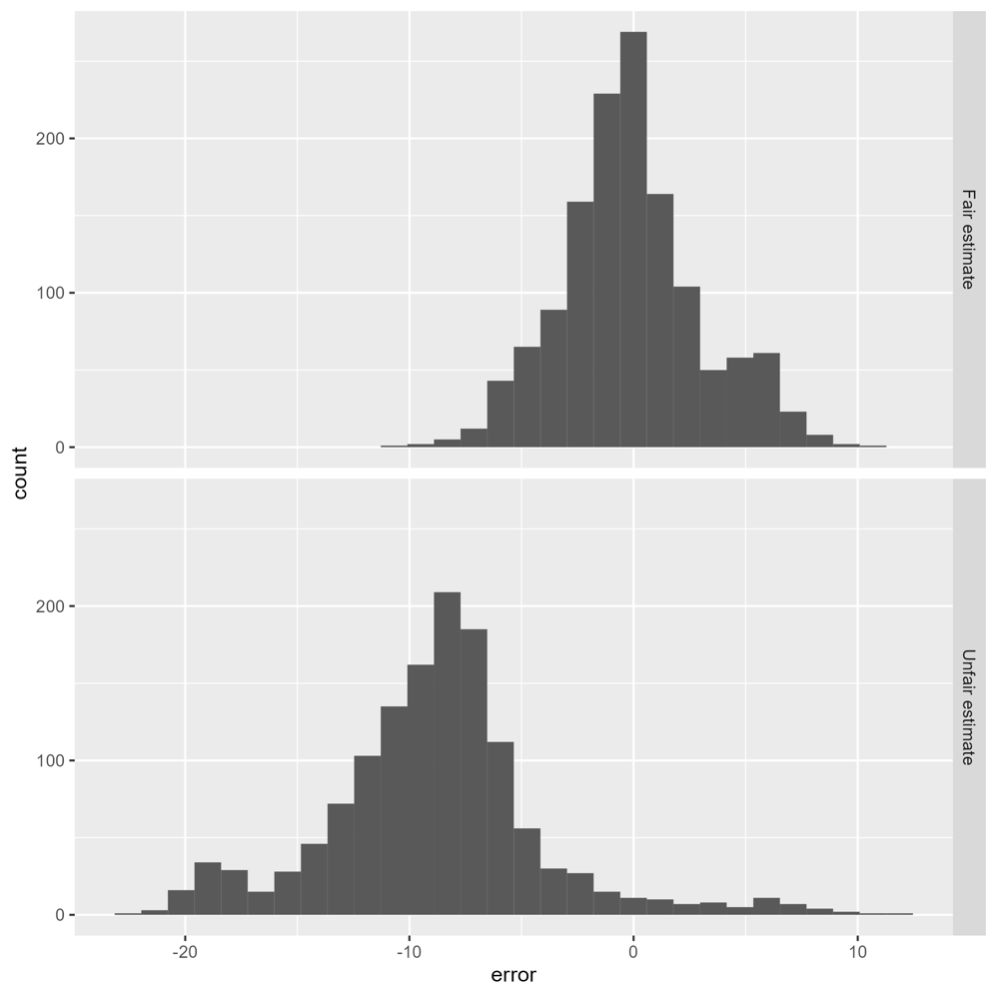}
  \caption{Counterfactual error for white women manipulated to be white men}
  \label{fig:fig6}
\end{figure}

\FloatBarrier
Figures \ref{fig:fig1}, \ref{fig:fig2},  and \ref{fig:fig3} show as expected that predictions differ much less across sensitive characteristics after pre-processing with DML. This has achieved approximate group-level equality in outcomes including across interactions between variables like race and gender as shown in Figure \ref{fig:fig4}.
Showing group level equalisation though is not the main goal of counterfactual fairness. It is an individual-level measure so should be judged on the basis of discrimination between real and counterfactual estimates. While it may make sense to compare all cases against the base case or some other yardstick case, a better approach is to break our sample up into subgroups with a shared expected outcomes before comparing to a base case. This is because the subgroup where expected outcome is lower (and sample size is also lower) might be expected to naturally have more noise than subgroups with expected outcome closer to the base case and a higher sample size. To get an indication of performance of this algorithm we look at two subgroups\textemdash  cases of non-white, non-men and cases of white women. In both cases, we select these subgroups out of the test sample and then regenerate a copy of the data with the same latent values but with gender and race characteristics changed so the case is now a white man. Figure \ref{fig:fig5} and Figure \ref{fig:fig6} below compare the difference between real and counterfactual cases (what we will call counterfactual error) for a DML Fairness estimate and an estimate using a fairness through unawareness approach \citep{dwork_fairness_2012} (i.e. a model where sensitive variables are simply excluded when the model is fit).

On the subgroup of non-white, non-men, we see that the fair estimate centres the error distribution with a mean error of 0.41 and a standard deviation of 3.68. The fairness through unawareness error has a mean on -12.74 and a standard deviation of 7.69. On both bias and variance metrics then, the DML fairness approach gives better performance than fairness through unawareness. In particular, the fact that the DML fairness error is essentially unbiased is encouraging. There may be some error, but it is equally likely to disadvantage a white man or a non-white, non-man. Of course actual performance may be worse on real data where the DGP is simpler and our assumptions of additive separability may not hold. In addition, given the consistency of DML, we would expect this error to shrink as sample size increases.

On the subgroup of white women, we see similar results with the DML fairness estimator giving a mean error of -0.10 and a standard deviation of 3.15 compared to the fairness through unawareness estimator which gives a mean error of -8.90 and standard deviation of 4.86. Here the drivers of discrimination are less complex than in the previous case as there are no race effects or interaction effects between race and gender.

\section{Application}
\label{sec:application}
As an application we test DML fairness on an example previously used by \citet{komiyama_nonconvex_2018}. The problem is a slightly bizarre one, predicting undergraduate GPA from the characteristics of law school students protecting age, gender, and race as sensitive variables. The data is drawn from the Fairml R package \citep{scutari_fairml_2022}. The list below (reproduced from the package documentation) shows all the variables used in this analysis. ugpa is used as the target variable with gender and race1 as sensitive variables. All other variables are taken as non-sensitive predictors. 
\begin{itemize}
\item age, an ordinal variable containing the student's age in years;
\item decile1, an ordinal variable containing the student's decile in the school given his grades in Year 1;
\item decile3, an ordinal variable containing the student's decile in the school given his grades in Year 3;
\item fam\_inc, an ordinal variable containing student's family income bracket (from 1 to 5);
\item lsat, a continuous variable containing the student's LSAT score;
\item ugpa, a continuous variable containing the student's undergraduate GPA;
\item gender, a nominal with levels "female" and "male";
\item race1, a nominal with levels "asian", "black", "hisp", "other" and "white";
\item cluster, an ordinal variable with levels "1", "2", "3", "4", "5" and "6" encoding the tiers of law school prestige;
\item fulltime, a binary variable with levels "FALSE" and "TRUE", whether the student will work full-time or part-time;
\item bar, a binary variable with levels "FALSE" and "TRUE", whether the student passed the bar exam on the first try.
\end{itemize}

\begin{figure}[!htp] 
  \includegraphics[scale = 0.6]{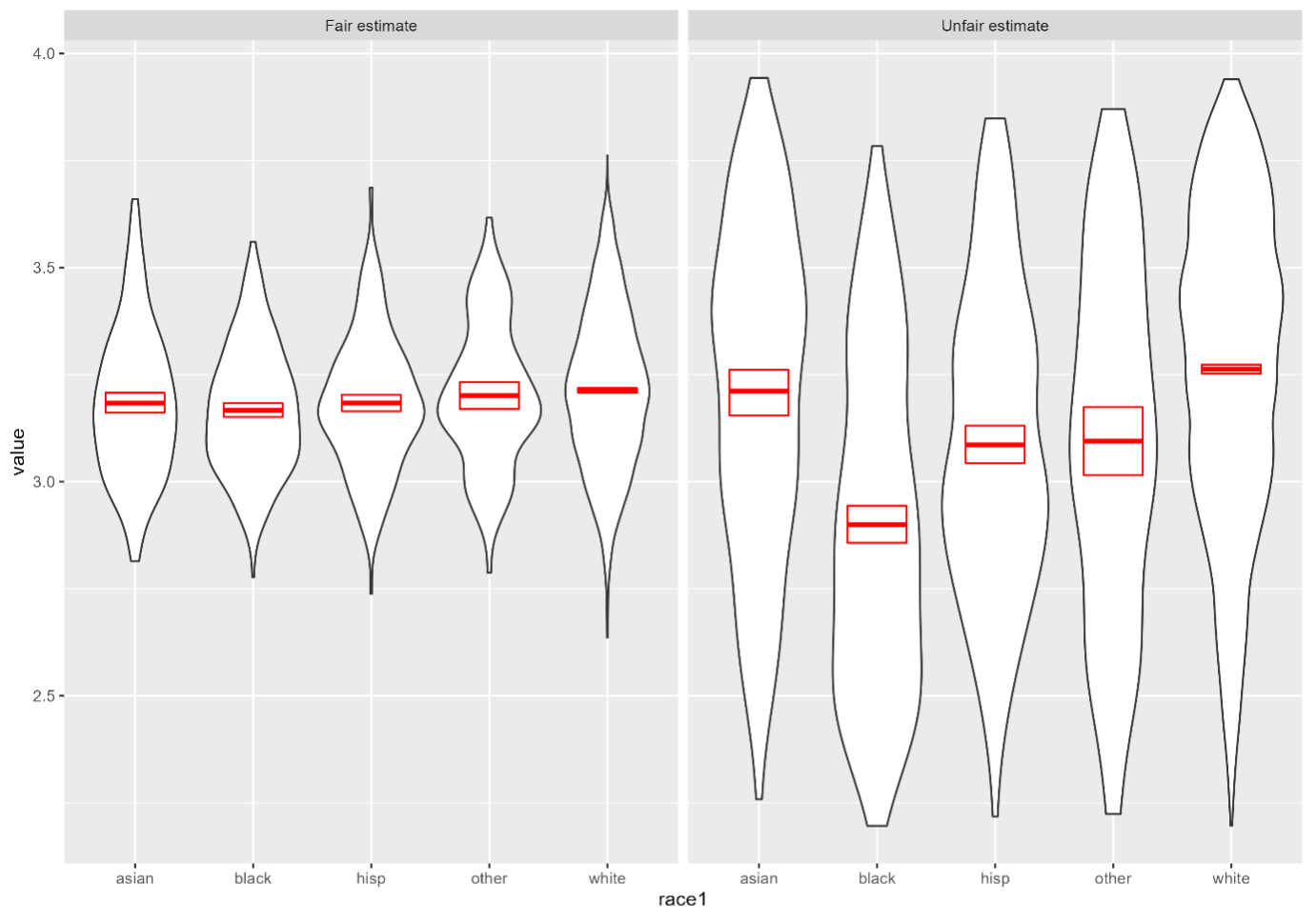}
  \caption{DML fairness compared with fairness through unawareness comparing GPA predictions by race}
  \label{fig:fig7}
\end{figure}

 \begin{figure}[!htp] 
  \includegraphics[scale = 0.6]{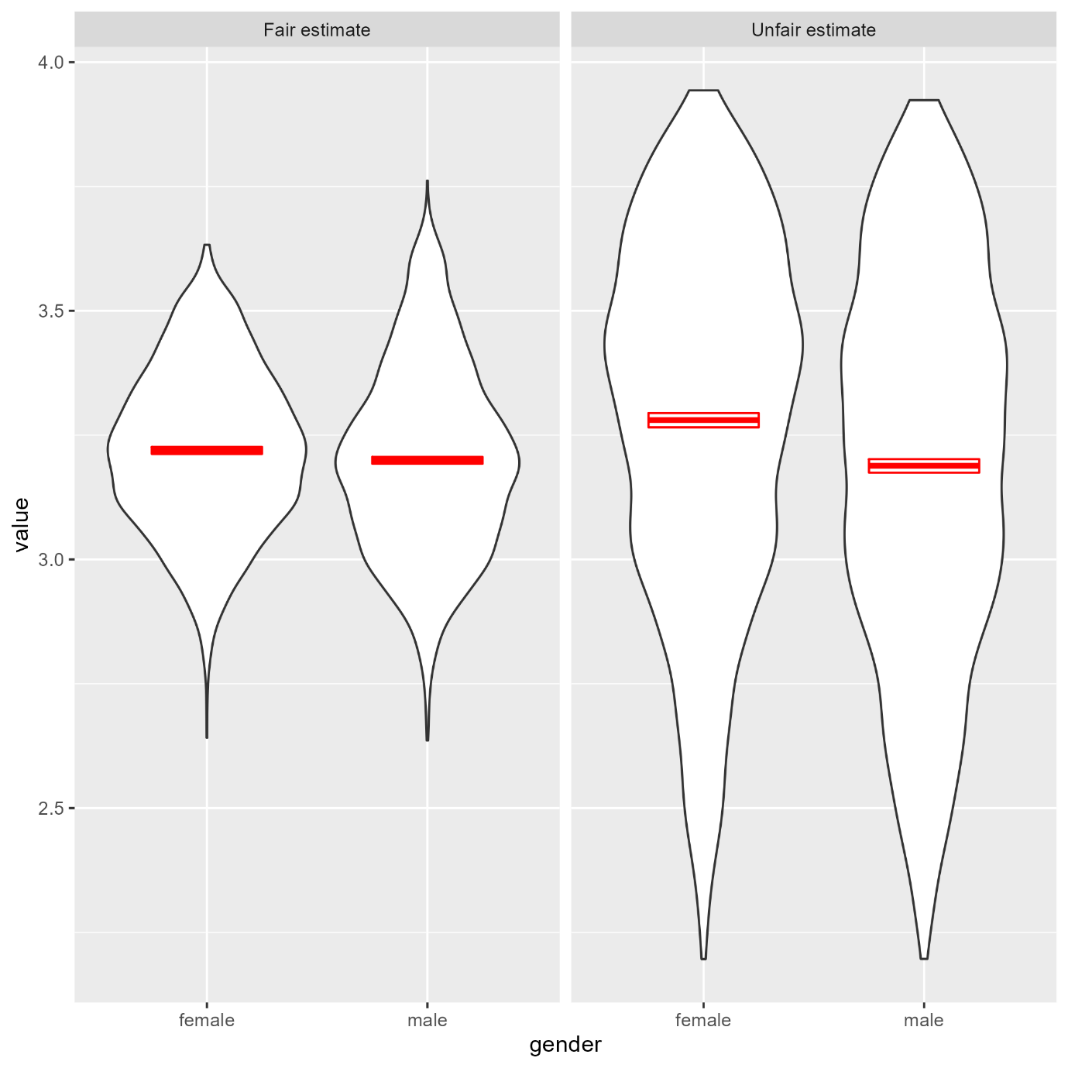}
  \caption{DML fairness compared with fairness through unawareness comparing GPA predictions by gender}
  \label{fig:fig8}
\end{figure}

 \begin{figure}[!htp] 
  \includegraphics[scale = 0.6]{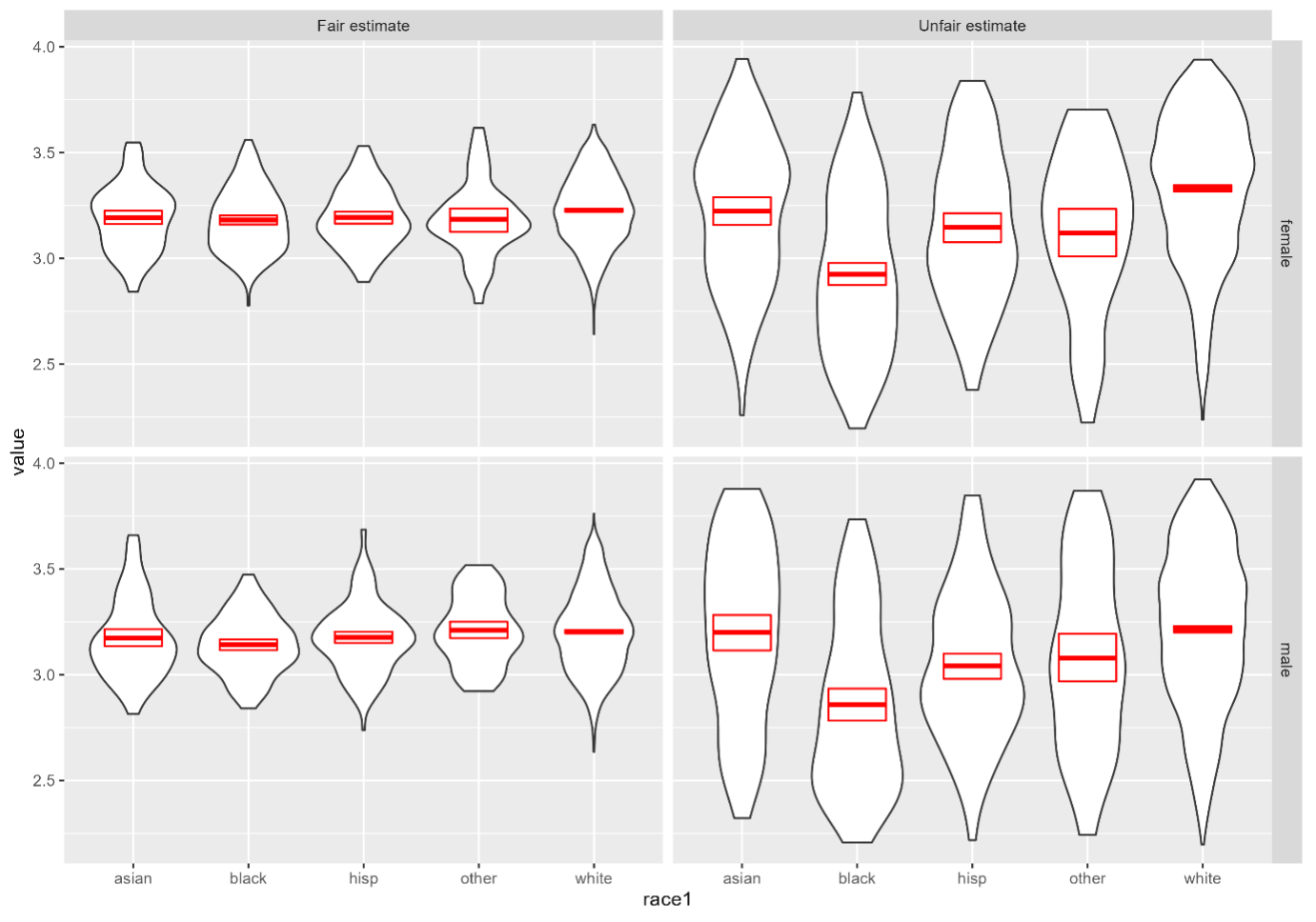}
  \caption{DML fairness compared with fairness through unawareness comparing GPA predictions by gender and race}
  \label{fig:fig9}
\end{figure}

\begin{figure}[!htp] 
  \includegraphics[scale = 0.35]{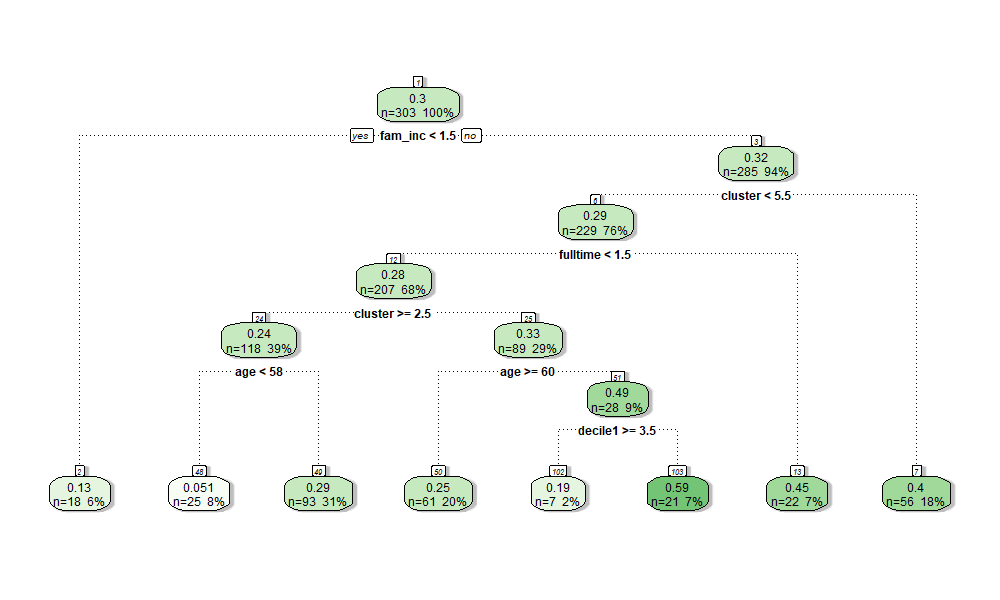}
  \caption{Adjustments for black students}
  \label{fig:fig10}
\end{figure}

Figures \ref{fig:fig7}, \ref{fig:fig8}, and \ref{fig:fig9} show the effect that DML fairness has on group-level fairness. Essentially we see that DML fairness is equalising means across sensitive variables and also reducing variance in estimates just as in the simulation study (this drop in variance is due to the replacement of variance from D with a constant from the base case estimate). 
Unfortunately, we cannot accurately estimate counterfactual error here. The reason for this is that due to the Fundamental Problem of Causal Inference, we cannot recover ground-truth counterfactuals (Holland 1986). One way we can try to understand what is happening here though is to look at who is benefiting most and suffering most from the use of DML Fairness adjustment. A decision tree can provide a simple explanation for the functioning of a black box model by fitting estimated outcomes from the model on the variables used to train it \citep{domingos_knowledge_1997}. In this case the tree is fit on the difference in outcomes between unawareness and DML Fairness models. It uses the evtree \citep{grubinger_evtree_2014} package to fit an optimal tree to a maximum depth of six splits. Figure \ref{fig:fig10} presents the results for black students. It shows how the adjustment varies by other characteristics with older students, part-time students, and students from poorer families benefiting more than others because these are characteristics that are strongly predicted by being black. (Note the age variable is the age of the former students at the time the data was collected, not the age of the students when they were at law school, this is why they seem unusually high). All plots by race and gender can be found in the Appendix.

\section{The limits of double machine learning for counterfactual fairness}
\label{sec:limits}
The major limitation of the DML Fairness approach is in how it conceptualises of counterfactuals. Counterfactual fairness is a very attractive approach as it offers the promise of erasing discrimination at the level of the individual and doing so with the tools of statistical modelling which machine learning practitioners already understand well. However, counterfactual reasoning is exceedingly complex, especially around complex social constructs. While estimation of outcomes with a given set of data can be done by machine learning algorithms, the actual counterfactual reasoning has to be done (or ignored) by human beings.
Many of the problems of AI fairness and ethics more broadly can be essentially seen as a failure to recognise the limitations of our computer systems. It makes sense that computer scientists and engineers are drawn to engineering solutions, but (at least for the foreseeable future) there are many domains of human decision-making that have no place (or at least a limited place) for AI systems. \citet{weizenbaum_computer_1976} identifies two types of activity that are often conflated into one in AI – deciding and choosing. Deciding is an optimisation problem, it is simply about finding the best option where there is a single knowable and exhaustive measure of net-benefit. Choosing on the other hand involves trade-offs where options cannot be compared as easily. By employing a complex counterfactual measure, we risk shifting focus to how to best estimate causal effects, rather than asking the values question of whether these sensitive constructs should be counterfactually manipulated and – if there continues to be inequitable outcomes after counterfactual adjustment on the variables – whether we should be comfortable with inequitable treatment with regard to a protected attribute even if a machine learning model says this is optimal. The rest of this section explores the limits of DML Fairness by seeking to answer four questions that threaten the validity or usefulness of the approach.
\subsection{Can we reason counterfactually around factors like race or gender?}
Often sensitive constructs are very theoretically complex and interact with other measures in ways that are difficult to model. After all, these often reflect entrenched marginalisations whose nature is largely socially constructed (and potentially reinforced by machine learning systems \citep{damour_fairness_2020}). This makes fairness both a ‘moving target’ and also something that might not be easily achieved just through imagining a manipulation of a variable. An example from \citet{hanna_towards_2020} examines the construct of race which is not only socially constructed but also to some extent constructed for the very purpose of discrimination. The authors argue that any use of race as a protected variable must be preceded by a process of deconstructing and questioning these categories in order for protection of a variable in a machine learning model to actually protect marginalised groups in AI decision-making. \citet{kasirzadeh_use_2021} apply this kind of critique to problematise counterfactual AI fairness and suggest a series of 10 assumptions that need to be validated with human judgement before a counterfactual approach is used to protect a construct.

Our variables are still only as valid as their measurement so we need to understand the complexities of doing research around complex protected attributes. As poor measurement only leads to attenuation of effect, it will also bias adjustment towards under-compensating for a construct \citep{Greene2003Econometric}. In addition, we need to be clear exactly what we want to be adjusting for and what counterfactual we are imagining exactly \citep{kohler-hausmann_eddie_2019}. For example, a complex question in economics is the measurement of the gender pay gap \citep{bishu_systematic_2017,kunze_gender_2008}. It is actually easy to measure the gender pay gap, it is hard to measure the causal relationship of gender on pay for the reason that a lot of factors affect pay and it is difficult to say what factors are and aren’t gendered. The pay gap is not just a product of gender directly (i.e. an employer making a biased decision when it comes to setting a woman’s pay), it is also a function of the kinds of jobs women are socialised into, differential care-giving responsibilities, gendered expectations around assertiveness in the workplace among many other factors \citep{bishu_systematic_2017,kunze_gender_2008}. The problem is that it is hard to know what variables we should actually be manipulating in order to draw counterfactuals. Are we imagining the employee is a man \textit{ceteris parabus}? Are we imagining a world where there is no construct of gender to affect employer decisions? Are we imagining a world without gendered choice of occupation? This is not a statistical problem, it is a problem of knowing exactly what the thing we’re trying to protect is and what it means to manipulate it to form a counterfactual \citep{pearl_causality_2009,kohler-hausmann_eddie_2019}. It is also a problem of judging whether we can adequately model this effect under an additive separability assumption. Returning to the language of Weizenbaum, while the fitting of a regression model is a decision with a clear loss function, this definition of the what the regression is actually measuring (i.e. the constructs we wish to protect and measurement of them) is a choice.
\subsection{Does it matter that counterfactual approaches are difficult to evaluate?}
Complex causal inference methods may also cause a problem because they admit less criticism than some of the simpler group-level fairness metrics. With group-level fairness metrics it is very easy to see whether or not the measures we have taken to address unfairness are working or not. It is very easy for example to say two groups have the same average prediction, it is much more difficult to say that the two groups’ predictions now reflect what they would be in a counterfactual world without race or gender differences. As already mentioned, counterfactual fairness can only be shown with the tools of causal inference and unfortunately due to the Fundamental Problem of Causal Inference, it is very difficult to determine whether something is fair or not. Of course critics could use the strategy of the ProPublica journalists who uncovered discrimination in the supposedly ‘race-blind’ COMPAS algorithm used to judge the risk of reoffending in bail and sentencing decisions in several US jurisdictions. They looked simply at large differences in outcomes for white and black defendants to shine a light on unfairness. But surely it is better to catch this problem before it becomes so grossly unfair that it is obvious even to outside observers. We can add some transparency through the use of Explainable AI systems tools \citep{lipton_mythos_2018,gunning_darpas_2019} or Interpretable AI models (which are ‘white-box’ models like decision trees where it is easy to see how a model is reaching its conclusions) \citep{rudin_stop_2019}. We can also establish confidence intervals around estimates through bootstrapping if this is computationally feasible. This would give some sense of how confident the user should be in the counterfactual point estimate. 

These technical solutions may help to evaluate whether we should trust a DML Fairness estimate or not, but they still leave open the question of whether counterfactually fair outcomes should even     be considered fair at all.

\subsection{Is adjusting within groups fair?}
One aspect of this method that is likely to be controversial is the reduction of variance within groups that can be seen in both the simulation and application. This amounts to penalising members of a marginalised group who have the characteristics of the advantaged group. For example, consider a poorer black law student who went to an overwhelmingly white elite law school and had to work harder and sacrifice more than their white classmates to get there. Is it 'fair' to just partial out this effect? Is it fair to treat them the same or differently from a black classmate from a wealthy family?
What within-group variation do we deem to be unfair and what variation do we deem to be legitimate in predicting outcomes? This relates to wickedly complex philosophical questions about social justice that are well beyond the scope of this paper \citep{sep-equal-opportunity}. For the purposes of demonstration we have hypothesised a latent driver of difference in outcomes and tried to strip away all factors besides this. However, a more nuanced view could be taken by modelling out the effect of sensitive variables through their effects on mediators. For example, one might partial out the effect of race via family wealth on choice of law school (given racial wealth disparities likely play a role in school selection). In addition, using orthogonalisation as a regulariser might be useful for allowing greater variation within categories. However, it would be important to carefully assess whether this was working as intended in each application.
\subsection{Are counterfactuals enough to achieve fairness?}
Achieving fair outcomes with DML Fairness is not as simple as plugging data into the algorithm and nominating the sensitive variables to protect. While counterfactual fairness might be useful for regression tasks, there may be other factors that should be taken into account in making in choosing (in the Weizembaumian sense) an outcome informed by that regression. There is a lot of complex social data that cannot be incorporated into a fairness algorithm. For example, note in Figure \ref{fig:fig7} that the fairness adjustment reduces the estimates for women’s GPAs relative to those of men. Men outnumbered women in the sample and historically, men have been the majority in law schools and the legal profession in the United States \citep{czapanskiy_women_1989}. Depending on what decision this regression is being used to inform, it is worth asking whether it is fair to lower women’s estimated GPAs in making this decision just because those in law school tend to have had higher GPAs. This is not an easy question to answer. In fact, a close parallel – the debate about affirmative action making it harder for Asian-Americans to gain admission to elite American universities – shows just how fraught value-judgements about this kind of fairness can be \citep{lee_asian_2021}. The question is ultimately not one of what quantitative adjustments should be made, but how we should understand the effects of sensitive variables on other variables and how we should understand ‘fair’ decisions in the context of a history of discrimination.

One solution to this problem is to use the regular DML fairness estimate as a kind of 'floor' that can be used where the estimate is more advantageous. This approach uses a decision rule that takes the maximum of the DML Fairness and fairness through unawareness estimates i.e. assuming higher scores are better
$$Y_{i,fair}=max\left(Y_{i,DMLfair},Y_{i,unaware}\right)$$
However, this creates another problem. As DML Fairness reduces estimate variance due to substituting the noisy function $\widehat{g_Y\left(D_i\right)}$ with the constant $\widehat{g_Y\left(D_{BC}\right)}$ meaning that if the maximum is used, all this will do is create a floor for values at $f ({\widetilde{X}}_i)+\widehat{g_Y}\left(D_{BC}\right)$ which of course still benefits groups with higher fairness through unawareness scores, but it privileges higher variance estimates in a way that might be undesirable. An alternative is to instead estimate group-specific base cases and allow these to be used where beneficial i.e.
$$Y_{i,fair}=max\left(Y_{i,DMLfair},Y_{i,groupBC}\right)$$
However, using an estimate like this cannot be an automatic process. There is judgement involved in setting the base case and there should be judgement involved in deciding when group-specific base-cases should be used and what these groups should be. 
There are also cases where specific positive discrimination policies might be fairer than an actual counterfactual approach. An example of this is where societies owe a kind of debt to a marginalised group due to past injustices for example indigenous peoples in settler states. Here, we might want to look beyond strictly counterfactual fairness in making decisions and make Weizenbaumian choices instead. Although it is possible to view these issues as issues of counterfactual fairness (that is positive discrimination should close the gap between ones outcome in the real-world and the outcome in a world where one’s ancestors were not enslaved, dispossessed of their land or otherwise victimised \citep{boxill_black_2022}), it is not feasible to solve this problem computationally. Counterfactual fairness may still prove useful for these problems, but it is not a simple fix. These issues call for value judgements in order to get around limits in the ‘imagination’ of a statistical counterfactual model.

\section{Conclusion}
In this paper we have provided an algorithm for pre-processing data to fit a predictive model that protects certain sensitive variables according to the counterfactual criterion for AI fairness in regression applications. It allows for the use of arbitrary machine learning methods both in the orthogonalisation stage for fitting nuisance functions and in the fitting of the actual predictive model. However, I have also tried to lay out why this solution for counterfactual fairness with minimal assumptions may be too good to be true. As with all approaches to fairness this one has its problems and the complexity and difficulty in measuring individual-level error means that it must be applied very carefully. We have recommended being very careful in defining protected variables, using interpretable / explainable models if possible and considering whether in some circumstances, outcomes under positive discrimination rules might actually be fairer than outcomes from DML Fairness.

This paper has assumed that the effect of sensitive variables is additively separable from other predictors, however, Chernozhukov et al. also consider a fully interactive model where this is not the case. Future work may generalise this approach to those kinds of cases. However, we are currently comfortable avoiding this as it means abandoning a guarantee of approximate group-level fairness and the ability to easily re-centre predictions. This is part of a larger theme this paper has tried to communicate. Thinking counterfactually requires not just fitting data, but challenging causal reasoning. In the case of counterfactual fairness where we are thinking about complex, socially fraught constructs this is particularly challenging. While DML Fairness can potentially address some AI fairness issues, if applied without care, it could also lead to harmful outcomes as fairness issues might continue to linger even after they are considered resolved. Finally, as regression is a relatively marginal field in machine learning, it would be valuable to develop a similar approach to the problem of classification.

\bibliographystyle{plainnat}
\bibliography{dmlfair}  

\newpage\section{Appendix: Tree plots summarising adjustments within each racial group and gender in application}\begin{figure}[!htp] 
  \includegraphics[scale = 0.35]{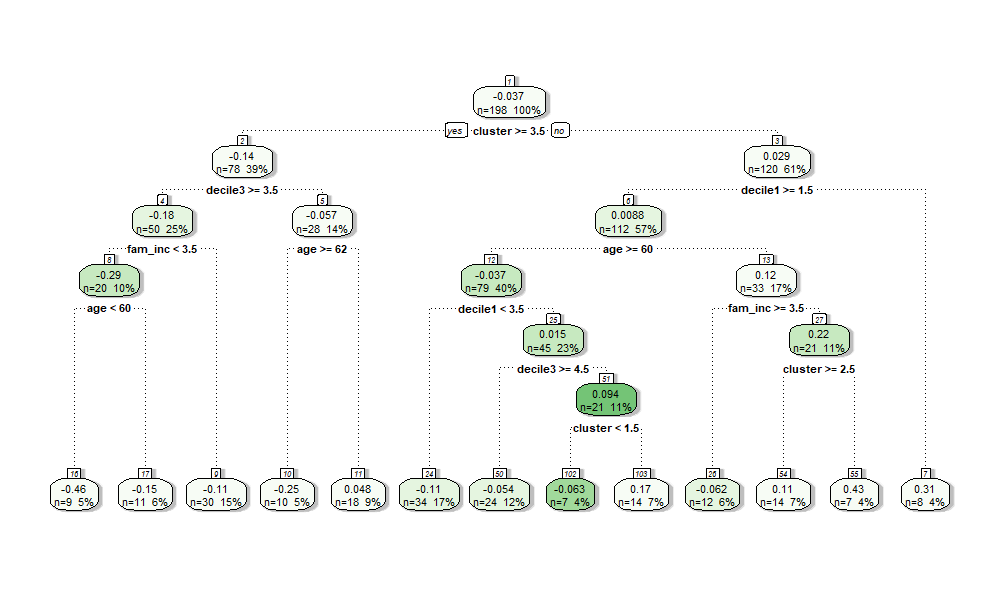}
  \caption{Adjustments for Asian students}
  \label{fig:fig11}
\end{figure}

 \begin{figure}[!htp] 
  \includegraphics[scale = 0.35]{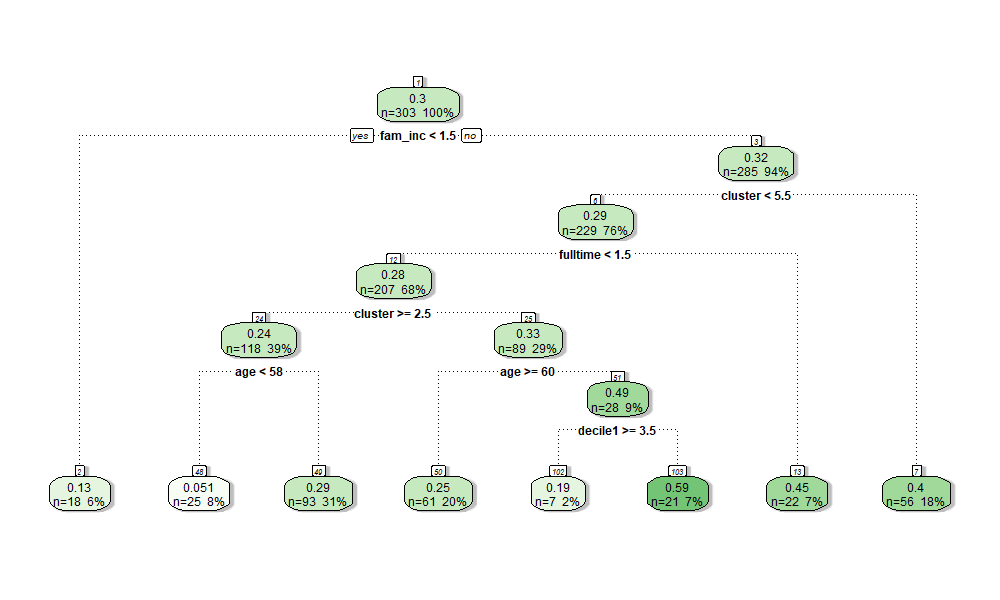}
  \caption{Adjustments for Black students}
  \label{fig:fig12}
\end{figure}

 \begin{figure}[!htp] 
  \includegraphics[scale = 0.35]{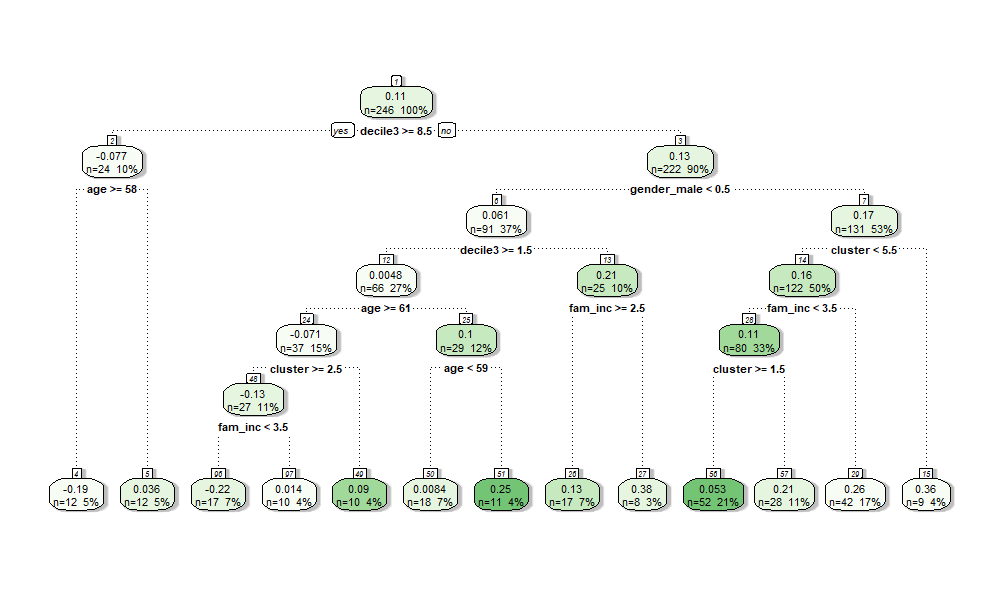}
  \caption{Adjustments for Hispanic students}
  \label{fig:fig13}
\end{figure}

 \begin{figure}[!htp] 
  \includegraphics[scale = 0.35]{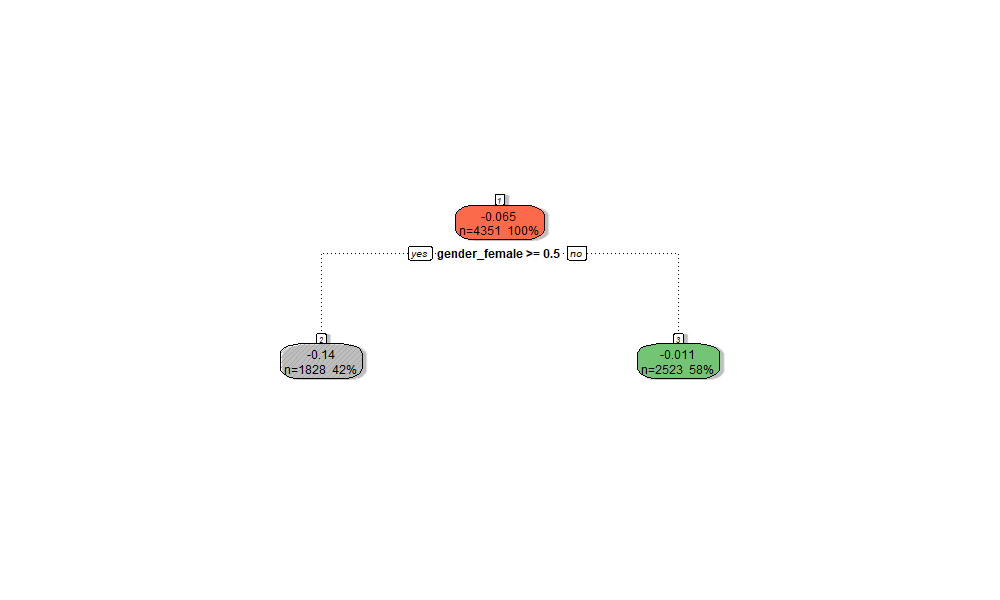}
  \caption{Adjustments for White students}
  \label{fig:fig15}
\end{figure}

 \begin{figure}[!htp] 
  \includegraphics[scale = 0.35]{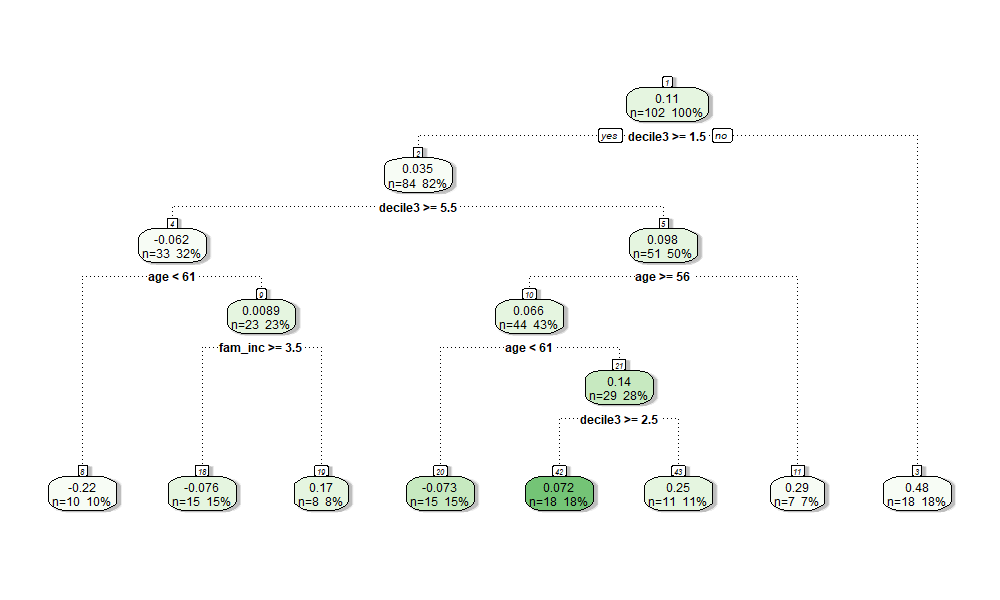}
  \caption{Adjustments for students of other races}
  \label{fig:fig14}
\end{figure}

 \begin{figure}[!htp] 
  \includegraphics[scale = 0.35]{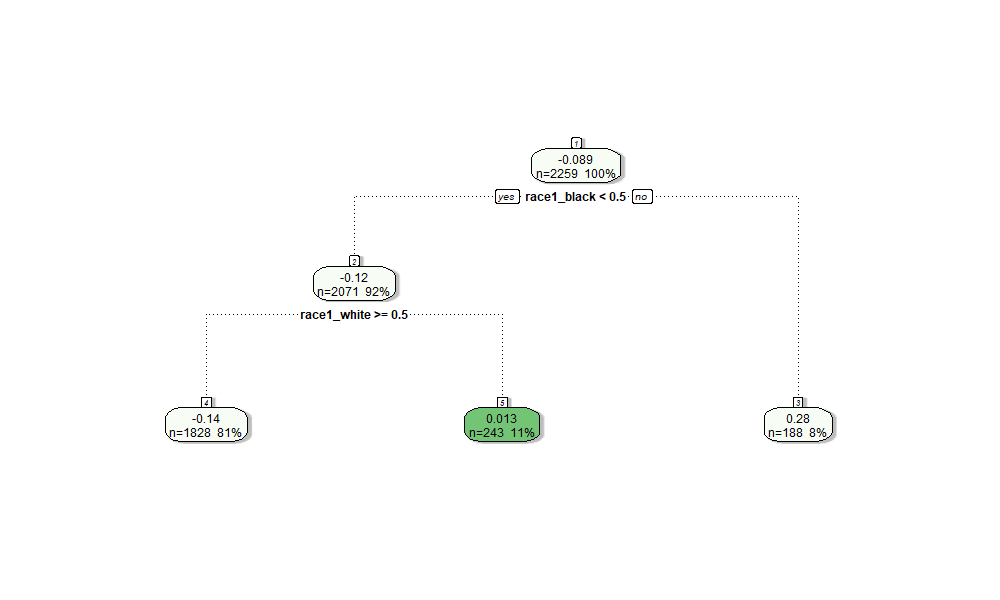}
  \caption{Adjustments for female students}
  \label{fig:fig16}
\end{figure}

 \begin{figure}[!htp] 
  \includegraphics[scale = 0.35]{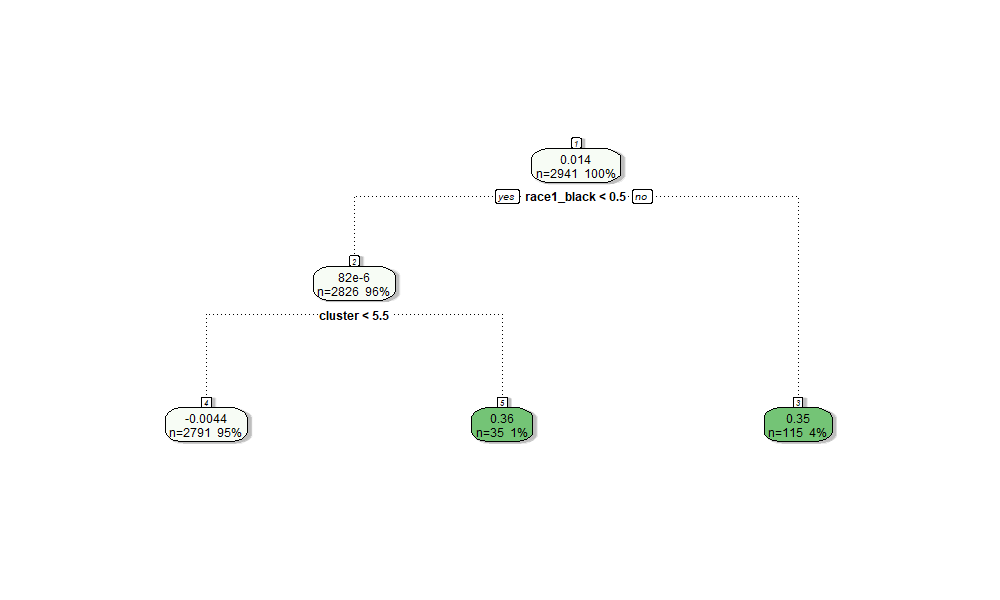}
  \caption{Adjustments for male students}
  \label{fig:fig16}
\end{figure}

\end{document}